\documentclass[journal, 10 pt, twoside]{ieeetran} 
\IEEEoverridecommandlockouts

\usepackage{cite}
\usepackage{amsmath,amssymb,amsfonts}
\usepackage{algorithm}
\usepackage[noend]{algpseudocode}
\usepackage{graphicx}
\usepackage{textcomp}
\usepackage{url}
\usepackage[caption=false, font=footnotesize]{subfig}
\usepackage{listings}
\usepackage{caption}

\usepackage{booktabs}
\usepackage{multirow}
\usepackage[table,xcdraw]{xcolor}
\usepackage{tabularx}

\definecolor{commentgreen}{rgb}{0,0.6,0}
\definecolor{keywordblue}{rgb}{0,0,0.8}
\definecolor{stringred}{rgb}{0.8,0,0}

\usepackage[htt]{hyphenat}

\lstdefinestyle{cpp}{
  backgroundcolor=\color{white},                   % Light background
  basicstyle=\ttfamily\fontsize{7.5pt}{6pt}\selectfont,            % Use the same font as text, footnote size
  numberstyle=\scriptsize\color{black},                   % Line numbers in gray and tiny
  numbersep=5pt,                                   % Space between line numbers and code
  numbers=left,                                    % Line numbers on the left
  frame=single,                                   % Single frame around the code
  frameround=tttt,
  rulecolor=\color{black},                         % Frame color
  captionpos=b,                                    % Caption position at the bottom
  breaklines=true,                                 % Automatically break long lines
  breakatwhitespace=true,                          % Break at whitespace
  stringstyle=\color{blue},                        % String style in blue for better readability
  keywordstyle=\color{black}\bfseries,           % Make keywords bold and purple
  commentstyle=\color{green!50!black}\itshape,   % Comments in green italics
  morekeywords={elif, if, else, while, break, const, },      % List of functions and keywords
}
\def\BibTeX{{\rm B\kern-.05em{\sc i\kern-.025em b}\kern-.08em
    T\kern-.1667em\lower.7ex\hbox{E}\kern-.125emX}}
\begin{document}

\urlstyle{tt}

\title{A ROS2 Interface for Universal Robots Collaborative Manipulators Based on ur\_rtde}

\author{\IEEEauthorblockN{Alessio Saccuti, Riccardo Monica, Jacopo Aleotti} \\
\IEEEauthorblockA{\textit{Department of Engineering and Architecture, University of Parma, Parma, Italy} \\
\{alessio.saccuti,riccardo.monica,jacopo.aleotti\}@unipr.it}}

\maketitle

\begin{abstract}
In this paper a novel ROS2 driver for UR robot manipulators is presented, based on the \texttt{ur\_rtde} C++ library.
The proposed driver aims to be a flexible solution, adaptable to a wide range of applications.
The driver exposes the high-level commands of Universal Robots URScripts, and custom commands can be added using a plugin system.
Several commands have been implemented, including motion execution along a waypoint-based path.
The driver is published as open source.
\end{abstract}
   

\section{Introduction}

Collaborative robots have become widely adopted due to their flexibility and safety in human-robot interactions.
Universal Robots (UR) is one of the leading manufacturers of collaborative manipulators.
UR robots can be programmed using URScript, a custom language developed by Universal Robots and executed on the robot controller.
URScript provides several high-level commands like, for example, linear motion to a point (\textit{moveL}).
UR robots can also be controlled remotely with various communication interfaces like the Real-Time Data Exchange interface (RTDE) \cite{rtde}, which allows external applications to communicate with the robot controller over a TCP/IP connection.
Real-time communication is achieved using a message exchange protocol where the robot sends its status at each control cycle (up to $500$ Hz) and the external application can change the value of robot controller registers.

As RTDE is a low level communication protocol, high-level software libraries have been developed to enable advanced robot commands, such as executing trajectories.
The official library published by Universal Robot supporting RTDE is \texttt{ur\_client\_library} \cite{ur_client_library}, which was used to develop the official ROS2 driver for UR robots: \texttt{ur\_robot\_driver} \cite{ur_ros2_driver}.
The \texttt{ur\_client\_library}, and in turn \texttt{ur\_robot\_driver}, offer a rather low level of abstraction where high-level URScript commands (like \textit{moveL}) are not exposed. 
Those commands are accessible only by sending custom URScript snippets to the robot.

This paper proposes \texttt{ur\_ros\_rtde}, a novel ROS 2 driver for Universal Robots.
Unlike the official ROS2 driver, the proposed solution is built on the \texttt{ur\_rtde} C++ library \cite{official_ur_rtde}, a RTDE interface that operates at a higher level of abstraction than \texttt{ur\_client\_library}.
By leveraging \texttt{ur\_rtde}, \texttt{ur\_ros\_rtde} exposes the high-level URScript commands as ROS2 action servers. 
Other commands have been implemented as a composition of URScript instructions.
Moreover, additional commands can be added to the driver, if needed, by leveraging a plugin system.
The proposed solution aims at providing a flexible driver, designed for easy adaptation to a wide range of applications, including academic research.

In summary, the contributions of this work are:

\begin{itemize}
\item a ROS2 interface for Universal Robot cobots based on \texttt{ur\_rtde} that exposes high-level URScript commands
\item the proposed driver is a more flexible alternative to \texttt{ur\_robot\_driver} thanks to a plugin system
\item the source code is published as open source\footnote{\texttt{https://github.com/SuperDiodo/ur\_ros\_rtde.git}}
\end{itemize}

\section{Related work}

The Universal Robots RTDE communication interface is well-known in literature and it was used in several works.
In \cite{rtde_1} and \cite{rtde_2} RTDE was adopted to control UR cobots.
In \cite{log_rtde_1}, \cite{log_rtde_2}, and \cite{log_rtde_3}, the RTDE interface was used only for data acquisition.

To facilitate the development of external applications for UR cobots, various higher-level software interfaces and drivers have been proposed based on RTDE.
In addition to the official software interface by Universal Robots (\texttt{ur\_client\_library}), a few alternatives have been developed by third-parties.
One of these software interfaces is \texttt{ur\_rtde} \cite{official_ur_rtde} by SDU Robotics, which was used in this work.
Another similar interface is \texttt{python-urx} \cite{python-urx}, which is a Python interface for tasks that do not require high control frequency.
A C\# software interface is provided in \textit{UnderAutomation} \cite{underautomation}.

To the best of our knowledge, the only ROS2 driver currently available is the official Universal Robot driver \texttt{ur\_robot\_driver}.
Unlike the driver proposed in this work, in \texttt{ur\_robot\_driver} most high-level URScript commands are accessible only through the low-level transmission of custom URScript snippets to the robot.
Also, there is no mechanism to verify the correctness of the snippet or to obtain feedback after execution.
Indeed, in previous works \cite{ros_driver_1}\cite{ros_driver_2}\cite{ros_driver_3}\cite{ros_driver_4}\cite{ros_driver_5} only default functionalities of the driver were used like accessing robot state or executing trajectories.
Another disadvantage of \texttt{ur\_robot\_driver} is that it requires a program running on the robot controller. 
In particular, the \textit{External Control URCap} package must be installed and running to control the robot.

In this work \texttt{ur\_rtde} was preferred over \texttt{ur\_client\_library} and \texttt{python-urx} due to its ease of use and the native availability of high-level URScript commands.
Moreover, the \texttt{ur\_rtde} software interface does not depend on the \textit{External Control URCap}, even if it is compatible with it.
The ease of use of \texttt{ur\_rtde} and its usefulness are evident given the number of works where it has been employed \cite{ur_rtde_1}\cite{ur_rtde_2}\cite{ur_rtde_3}\cite{ur_rtde_4}\cite{ur_rtde_5}\cite{ur_rtde_6}\cite{ur_rtde_7}.
However, no ROS2 interface for \texttt{ur\_rtde} is currently available, other than the one proposed in this paper.

\begin{figure}[t!]
  \centering
  \includegraphics[width=1.0\linewidth]{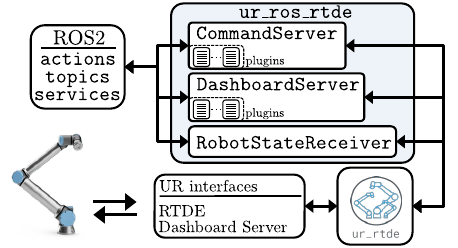}
  \caption{\label{fig:ur_ros_rtde} Diagram of \texttt{ur\_ros\_rtde}.}
\end{figure}

\section{Software description}

The proposed software architecture is summarized in Fig.~\ref{fig:ur_ros_rtde}.
The proposed driver includes three ROS2 nodes: \textit{RobotStateReceiver}, \textit{CommandServer}, and \textit{DashBoardServer}.
All three nodes (summarized in Section~\ref{sec:soft:rtde}) use the \texttt{ur\_rtde} interface for communication with the robot.
The \textit{RobotStateReceiver} node retrieves the robot state (Section~\ref{sec:soft:state}).
The \textit{CommandServer} and \textit{DashBoardServer} nodes expose ROS2 actions to send high-level commands to the robot, and they can be extended using the plugin system of \texttt{ur\_ros\_rtde} (Section~\ref{sec:soft:commands}).
More advanced plugins which extend the robot control script can also be created (Section~\ref{sec:soft:plugin}).

\subsection{Overview of \texttt{ur\_rtde}}
\label{sec:soft:rtde}

The \texttt{ur\_rtde} interface by SDU Robotics is a C++ library with Python bindings for controlling UR cobots over RTDE.
The library has a modular structure where each module (called ``interface'' by the developers) exposes a subset of the robot functionalities.
Four modules of \texttt{ur\_rtde} are used in this work: the \textit{RTDE Control} interface, which sends motion commands; the \textit{RTDE Receive} interface, which is used for retrieving the current robot state; the \textit{RTDE IO} interface, which handles the robot I/O pins; and the \textit{Dashboard Client} interface, which can be used to send commands to the teach pendant GUI remotely.
Each module is represented by a main class, which can be instantiated independently from the others.

Among the listed modules, only the \textit{RTDE Control} interface requires a control script running on the robot.
The control script is automatically uploaded to the robot controller when the \textit{RTDE Control} class is instantiated, without requiring any installation on the robot, unlike the official \texttt{ur\_robot\_driver}.
Nonetheless, the \textit{RTDE Control} interface is still compatible with the official \textit{External Control URCap} control script if it is already running on the robot.
To prevent conflicts, at any time only one instance of the \textit{RTDE Control} interface can be connected to the robot, unlike the other interfaces.

The RTDE protocol is a real-time protocol which operates on a TCP connection, synchronized with the internal robot control loop.
The RTDE protocol defines input and output registers, which correspond to variables residing in the robot controller memory.
Registers are used by the \textit{RTDE Control} interface for communication with the script running on the robot controller.
At each control loop, the current value of the output registers is sent by the robot controller to the connected client.
Some output registers have pre-defined meaning and they contain information about the current robot state (e.g., joint angles).
Other output registers can be configured, and can be used for custom communication from the script running on the controller to the client.
Similarly, the client can send messages through the TCP connection to change the value of the configurable input registers.

In order to execute a command on the robot, \textit{RTDE Control} sets a specific input register to the index of that command.
At each control loop iteration, in the \texttt{process\_cmd} function, the control script uses the value of the input register to select the command to be executed on that iteration.
Command parameters are passed using other input registers, and results are returned using output registers.

\subsection{Retrieving the robot state}
\label{sec:soft:state}

The ROS2 interface to access the current robot state is provided by the \textit{RobotStateReceiver} node of \texttt{ur\_ros\_rtde}.
The node was designed to depend only on the \textit{RTDE Receive} interface of \texttt{ur\_rtde}.
Hence, it does not require a control script running on the robot controller, and it does not interfere with other operations on the robot.
Any number of \textit{RobotStateReceiver} nodes may be running to monitor the robot state, potentially on different physical machines.

The \textit{RobotStateReceiver} node uses the \textit{RTDE Receive} interface to iteratively read the robot state with a configurable frequency.
The robot state includes data such as joint positions, forces and torques.
At each iteration, each component of the robot state is published to a ROS2 topic, in the standard format, if available.
Moreover, the last robot state is also stored in the node.
The stored robot state can be accessed at any time by calling a service (Table \ref{tab:information}).

\renewcommand{\arraystretch}{0.8}
\setlength{\tabcolsep}{4pt} 

\begin{table}[t]
\centering
\caption{\label{tab:information} Topics and services advertised by the \textit{RobotStateReceiver}.}
\resizebox{\columnwidth}{!}{
\begin{tabular}{@{}ccc@{}}
\toprule
\textbf{Type}      & \textbf{Topic}            & \textbf{Service}                    \\ \midrule
Joint state        & \texttt{/joint\_states}   & \texttt{/get\_joint\_state}         \\
TCP pose           & \texttt{/tcp\_pose}       & \texttt{/get\_tcp\_pose}            \\
Wrench             & \texttt{/wrench}          & \texttt{/get\_wrench}               \\
IO state     & \texttt{/io\_state}       & \texttt{/get\_io\_state}            \\ \bottomrule
\end{tabular}}
\end{table}

The joints' positions and velocities are published to the \texttt{joint\_states} topic, following the naming convention adopted in software such as MoveIt! \cite{moveit} and RViz \cite{rviz}.
The \texttt{tcp\_pose} topic contains the current pose of the tool contact point (TCP), which is defined using the teach pendant in the installation phase of the robot.
The \texttt{wrench} topic provides information about the force and the torque read by the robot force-torque sensor.
The \texttt{io\_state} topic reports the state of the robot and controller I/O digital pins.

\begin{figure}[t!]
  \centering
  \includegraphics[width=1.0\linewidth]{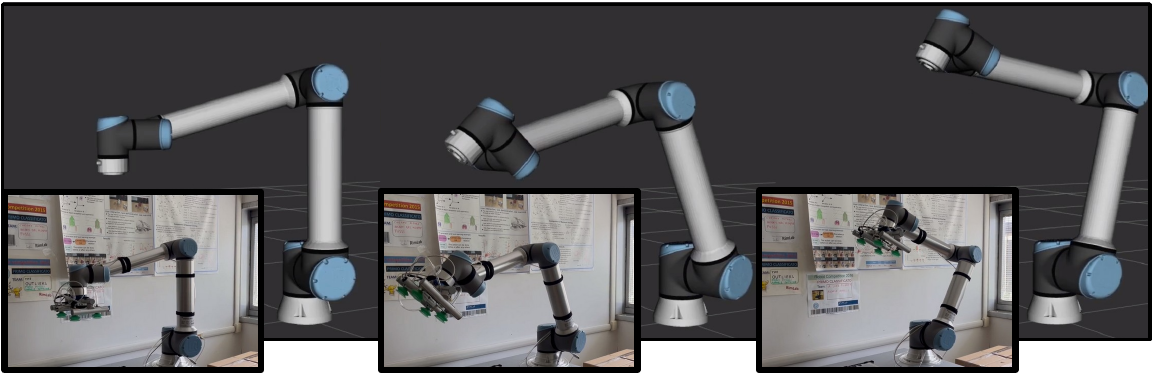}
  \caption{\label{fig:simulation} Digital twin of a UR cobot.}
\end{figure}

The \texttt{joint\_states} topic can be provided as input to a standard \texttt{robot\_state\_publisher} ROS2 node.
Hence, a digital twin simulation of the robot can be displayed in RViz (Figure \ref{fig:simulation}).
The \texttt{robot\_state\_publisher} node must be configured using the UR cobot URDF (Unified Robot Description Format) description.
However, unlike the official \texttt{ur\_robot\_driver}, \texttt{ur\_ros\_rtde} by itself does not depend directly on an URDF for accessing the robot state.
The \textit{RobotStateReceiver} provides a service for pausing the retrieval of the actual robot joints' positions to temporarily display a custom robot configuration.
In this case, the last joint positions are published to the \texttt{joint\_states} topic and the joints' positions can be updated externally by sending a message to the \texttt{fake\_joint\_states} topic.

\subsection{Robot commands and plugin system}
\label{sec:soft:commands}

The \textit{CommandServer} and \textit{DashBoardServer} nodes expose ROS2 actions to send high-level commands to the robot.
The \textit{CommandServer} is designed to implement high-level commands.
Instead, the \textit{DashBoardServer} implements commands which only use the \textit{Dashboard Client} interface.
The two nodes leverage a flexible plugin system where each plugin exposes different action servers.
All plugins are loaded using the ROS2 plugin system.
The plugins share the same instance of the \texttt{ur\_rtde} interface, hence only one connection (for each node) is created.

Three types of plugins were defined: the \textit{command} and \textit{extension} plugins extend the \textit{CommandServer} node, and the \textit{dashboard} plugins extends the \textit{DashBoardServer}.
Each plugin must implement the \textit{start\_action\_server} method, where the plugin should initialize itself, for example by creating an action server.
To speed-up the definition of many action servers, a simplified templated C++ action server implementation is also provided within \texttt{ur\_ros\_rtde}.
Plugins with the \textit{command} interface can access the common instance of the \textit{RTDE Receive}, \textit{RTDE Control}, \textit{RTDE IO} and \textit{Dashboard Client} interfaces of \texttt{ur\_rtde}.
The \textit{dashboard} plugins can only access the \textit{Dashboard Client} interface.
The \textit{extension} plugins extend the robot control script (Section~\ref{sec:soft:plugin}).

More than $40$ plugins have already been implemented in \texttt{ur\_ros\_rtde}.
Each plugin instantiates a single ROS2 action server, with a custom action type with the command input parameters and output values.
In most plugins, the action callback simply calls a high-level command by \texttt{ur\_rtde} and returns the results.
Examples are the \textit{moveJ} command, which allows linear movement of the robot in joint space, and the \textit{moveUntilContact} command, which moves the robot at constant speed until a collision is detected.

A notable plugin is \textit{ExecuteTrajectory}, which allows the execution of a motion along a path described by a sequence of waypoints.
To accomplish this, the \textit{ExecuteTrajectory} integrates a trajectory planner and a control loop based on the \textit{servoJ} function of \texttt{ur\_rtde}.
Execution of a motion given a path is not present in \texttt{ur\_robot\_driver}, where a time-parameterized trajectory must always be provided instead.

\subsection{Extending the \texttt{ur\_rtde} control script}
\label{sec:soft:plugin}

Plugins of the \textit{extension} type are used to extend the control script running on the robot, so that custom instructions can be called by \texttt{ur\_ros\_rtde}.
For example, add-on devices such as robot grippers often require software extensions, called URCaps, to be installed on the robot controller.
URCaps typically provide custom instructions which are not part of standard URScript.
The interface of \textit{extension}-type plugins requires the implementation of a method which returns the changes to be applied to the control script.
The changes should include a URScript code snippet with the custom instructions to be executed.
Also, the changes may include a preamble to be added to the URScript header.

The \texttt{ur\_ros\_rtde} \textit{CommandServer} assigns a non-zero unique ID to each \textit{extension} plugin.
The code snippet of each plugin is added in the control script \texttt{process\_cmd} function so that, when input register $18$ contains one of the unique IDs, the corresponding snippet is executed.
The input float register $18$ was chosen for this purpose, as it is not used by the unmodified \texttt{ur\_rtde} interface.
The output float register $18$ is used to signal the completion of the snippet execution.
Hence, C++ code in the plugin can trigger the snippet by setting input register $18$ to its own unique ID.
Then, the code can wait for end of execution by monitoring output register $18$.

\begin{figure}[t!]
  \centering
  \includegraphics[width=1\linewidth]{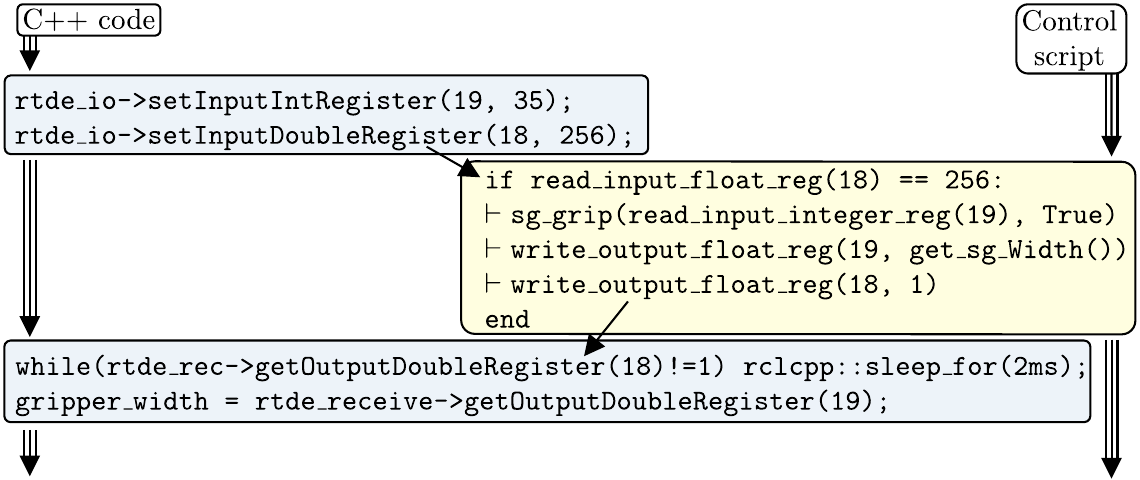}
  \caption{\label{fig:snippets} Snippet for OnRobot \textit{sg\_grip} method (yellow background) and corresponding C++ code (blue background).}
\end{figure}

Fig.~\ref{fig:snippets} illustrates an example URScript snippet to close an OnRobot soft gripper, and the corresponding C++ code to trigger the snippet.
The C++ code first sets the target gripper width in integer input register $19$, and then triggers the snippet by setting register $18$ to its unique ID ($256$ in this case).
The snippet uses the \textit{sg\_grip} command provided by the OnRobot URCap to close the gripper to the value in input register $19$.
After the execution of the command, the snippet stores the reached gripper width into output register $19$ and signals end of execution using output register $18$.
Hence, the C++ code can wait for end of execution using register $18$ and then retrieve the gripper width from register $19$.

\section{Example of a custom plugin}
\label{sec:examples}

\begin{figure}[t!]
  \centering
  \includegraphics[width=1\linewidth]{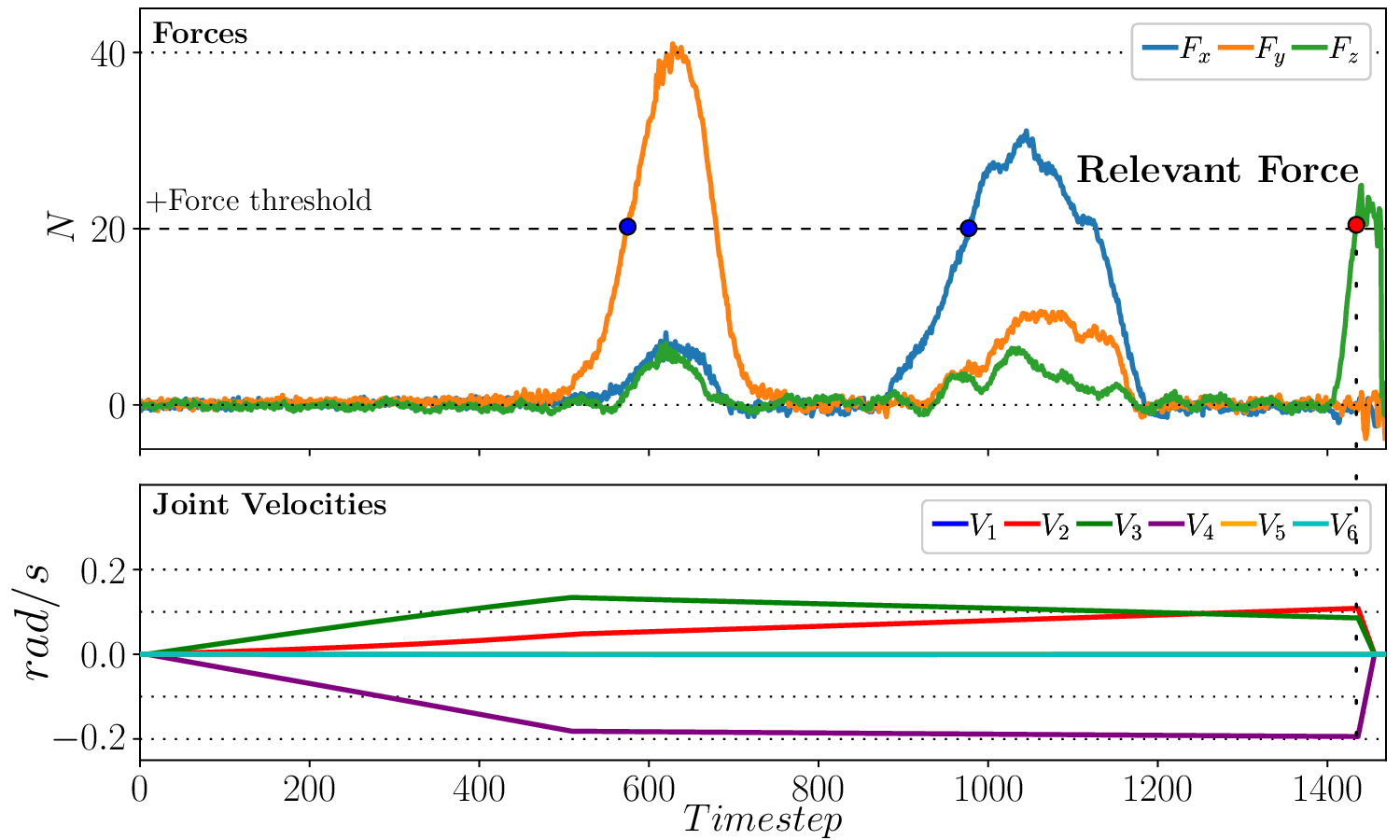}
  \caption{\label{fig:force_sensing} Forces (top) and joints' velocities (bottom) in the experiment with the \textit{MoveDownUntilForce} plugin. Blue and red dots highlight the instants in which the force exceeded the threshold. At the first two instants (blue dots) the force was ignored since the force direction is not along the flange Z-axis.}
\end{figure}

The plugin system of \texttt{ur\_ros\_rtde} can be leveraged to create custom commands for specific use cases.
As an example, the \textit{MoveDownUntilForce} plugin provides an action which moves the end-effector vertically until a force above a threshold is detected along the flange Z-axis.
The action is useful, for example, when placing an object on a surface.
The plugin overcomes a limitation of the similar \textit{moveUntilContact} function of \texttt{ur\_rtde}, where the threshold is not configurable.

Listing~\ref{lst:move_down_until_force_plugin} shows the content of the action callback.
First, the force-torque sensor is reset (Line~\ref{lst:move_down_until_force_plugin:ft_reset}). 
In Line~\ref{lst:move_down_until_force_plugin:pose}, the \textit{RTDE Receive} interface is used to obtain the robot current pose, which is then used to compute the target pose by adding a vertical offset of $1$ meter. 
The command \textit{moveL} (Line~\ref{lst:move_down_until_force_plugin:movel}) starts the linear movement of the robot to the target point. 
During the movement a loop retrieves the current force measured by the robot force-torque sensor (Lines~\ref{lst:move_down_until_force_plugin:while}-\ref{lst:move_down_until_force_plugin:while_end}). 
When a force of $20$ N is detected along the flange Z-axis (Line~\ref{lst:move_down_until_force_plugin:if}), the robot is stopped using the \textit{stopJ} command (Line~\ref{lst:move_down_until_force_plugin:stopj}).
The example is provided in the driver repository within the \texttt{ur\_ros\_rtde\_tutorials} package.
%Execution is also shown in the attached video.

\begin{center}
\begin{minipage}{0.81\linewidth}  % Adjust width as needed
\begin{lstlisting}[style=cpp, escapechar=|]
rtde_control->zeroFtSensor(); |\label{lst:move_down_until_force_plugin:ft_reset}|
auto goal_pose=rtde_rec->getActualTCPPose(); |\label{lst:move_down_until_force_plugin:pose}|
goal_pose[2] -= 1;
rtde_control->moveL(goal_pose,0.1,0.1,true);|\label{lst:move_down_until_force_plugin:movel}|
while (true) { |\label{lst:move_down_until_force_plugin:while}|
  auto f = rtde_rec->getActualTCPForce(); 
  if (abs(f[2]) > 20.0) {     |\label{lst:move_down_until_force_plugin:if}|
    rtde_control->stopJ(5.0); |\label{lst:move_down_until_force_plugin:stopj}|
    break; 
  } |\label{lst:move_down_until_force_plugin:while_end}|
  rclcpp::sleep_for(2ms);
}
\end{lstlisting}
\end{minipage}
\captionof{lstlisting}{Action callback of the \textit{MoveDownUntilForce} plugin.\label{lst:move_down_until_force_plugin} }
\end{center}

Fig.~\ref{fig:force_sensing} displays the force detected by the robot and the joints' velocities in a test experiment of the plugin.
As expected, the robot continued the motion until a force along the flange Z-axis exceeded the threshold.
Then, the joint velocities dropped to zero, stopping the robot.
Horizontal forces along the X and Y axes, applied by hand to interfere with the robot, were ignored.

For comparison, the same task was implemented in a separate ROS2 node using the \texttt{ur\_robot\_driver}. 
Due to the verbosity of ROS2, only a pseudocode of the implementation is shown in Listing~\ref{lst:move_until_force_ros}.
%The complete script can be found in the supplementary material.

At Line~\ref{lst:move_until_force_ros:reset_ft}, the force-torque sensor is reset by calling a service. 
At Lines~\ref{lst:move_until_force_ros:sub_ft} and \ref{lst:move_until_force_ros:sub_joint}, topics are subscribed to update the \texttt{force\_z} and \texttt{speed} variables, which represent the vertical force measured by the robot force-torque sensor and the actual joint speeds, respectively.
As the high-level \textit{moveL} and \textit{stopJ} commands are not available in \texttt{ur\_robot\_driver}, code snippets containing the commands are defined at Lines~\ref{lst:move_until_force_ros:snippet_start}–\ref{lst:move_until_force_ros:snippet_end}.
The robot starts moving at Line~\ref{lst:move_until_force_ros:move}, when the URScript snippet containing \textit{moveL} is sent. 
The process then waits until a vertical force of $20$ N is detected (Line~\ref{lst:move_until_force_ros:force_sensing}).
When the vertical force read by the robot exceeds the threshold the robot is stopped by sending the snippet containing the \textit{stopJ} command (Line~\ref{lst:move_until_force_ros:stop}). 
As there is no mechanism to wait for the end of a snippet, the node cannot determine when the stop command ends. 
Hence, at Line~\ref{lst:move_until_force_ros:wait_stop}, the process waits until the robot joint velocities drop below a given threshold. 
Finally, the \textit{External Control URCap} must be restarted (Line~\ref{lst:move_until_force_ros:restart}) due to a further limitation of \texttt{ur\_robot\_driver} where URScript snippets stop programs running on the robot controller.

\begin{center}
\begin{minipage}{0.86\linewidth}  % Adjust width as needed
\begin{lstlisting}[style=cpp, escapechar=|]	
force_z = 0, speed = 0
call_service("reset_ft_service"); |\label{lst:move_until_force_ros:reset_ft}|
subscribe("ft_topic",{force_z=get_force_z();}); |\label{lst:move_until_force_ros:sub_ft}|
subscribe("joint_states",{speed=get_speed();}); |\label{lst:move_until_force_ros:sub_joint}|
move_script = "def move_script(): |\label{lst:move_until_force_ros:snippet_start}|
    		 pose=get_actual_tcp_pose()
    		 pose[2]=pose[2]-1
    		 movel(pose,a=0.1,v=0.1,r=0)
    	       end";
stop_script = "def script_test(): |\label{lst:move_until_force_ros:snippet_end}|
    		 stopj(a=5.0)
    	       end"; 
publish("uscript_topic", move_script); |\label{lst:move_until_force_ros:move}|
while (abs(force_z) < 20.0) { //wait }   |\label{lst:move_until_force_ros:force_sensing}|
publish("uscript_topic", stop_script); |\label{lst:move_until_force_ros:stop}|
while (abs(speed) > 0.005) { //wait } |\label{lst:move_until_force_ros:wait_stop}|
call_service("restart_program"); |\label{lst:move_until_force_ros:restart}|
\end{lstlisting}
\end{minipage}
\captionof{lstlisting}{Implementation of a custom command similar to \textit{MoveDownUntilForce}, using \texttt{ur\_robot\_driver}.\label{lst:move_until_force_ros} }
\end{center}

The implementation of Listing~\ref{lst:move_until_force_ros} shows that several workarounds are needed to use the \texttt{ur\_robot\_driver}.
Indeed, high-level commands are accessible only through URScript snippets and also do not provide feedback after execution.
Instead, a plugin on \texttt{ur\_ros\_rtde} can access the high-level commands of the \texttt{ur\_rtde} directly.
Moreover, access to \texttt{ur\_robot\_driver} is done through messages and services, which have higher overhead than the \texttt{ur\_rtde} interface.
Hence, the implementation of critical control tasks as \texttt{ur\_ros\_rtde} plugins may be preferable.

\section{Conclusion}
This paper presented \texttt{ur\_ros\_rtde}, a novel ROS2 driver for Universal Robots.
Unlike the official \texttt{ur\_robot\_driver}, the proposed driver exposes high-level URScript commands, and custom commands can be added by leveraging a plugin system.
The driver is published as open source, with several plugins already implemented, including motion along a sequence of waypoints.
A comparison against the implementation of custom commands using \texttt{ur\_robot\_driver} was also reported.

\bibliographystyle{IEEEtran}
\bibliography{biblio}

@software{moveit,
  author = {{Ioan A. Sucan and Sachin Chitta}},
  title = {MoveIt},
  url = {moveit.ros.org}
}

@article{rviz,
  title={RViz: a toolkit for real domain data visualization},
  author={Hyeong Ryeol Kam and Sung-Ho Lee and Taejung Park and Chang-Hun Kim},
  journal={Telecommunication Systems},
  year={2015},
  volume={60},
  pages={337-345}
}

@software{rtde,
  title        = {RTDE},
  author       = {Universal Robot},
  note         = {\url{https://www.universal-robots.com/articles/ur/interface-communication/real-time-data-exchange-rtde-guide}}
}

@INPROCEEDINGS{official_ur_rtde,
  author={Lindvig, Anders Prier and Iturrate, I{\~n}igo and Kindler, Uwe and Sloth, Christoffer},
  booktitle={IEEE/SICE International Symposium on System Integration (SII)}, 
  title={ur\_rtde: An Interface for Controlling Universal Robots (UR) using the Real-Time Data Exchange (RTDE)}, 
  year={2025},
  volume={},
  number={},
  pages={1118-1123},
  doi={10.1109/SII59315.2025.10871000}}

@software{ur_ros2_driver,
  title        = {ur\_ros2\_driver},
  author       = {Universal Robots},
  note         = {\url{https://github.com/UniversalRobots/Universal_Robots_ROS2_Driver}}
}

@software{ur_client_library,
  title        = {ur\_client\_library},
  author       = {Universal Robots},
  note         = {\url{https://github.com/UniversalRobots/Universal_Robots_Client_Library}}
}

@software{underautomation,
  title        = {UnderAutomation},
  author       = {UnderAutomation},
  note         = {\url{https://github.com/underautomation/UniversalRobots.NET}}
}

@software{python-urx,
  title        = {python-urx},
  author       = {Olivier Roulet-Dubonnet },
  note         = {\url{https://github.com/SintefManufacturing/python-urx}}
}

@INPROCEEDINGS{rtde_1,
  author={Zakaria, Ungku M.Z. Ungku and Mustaza, Seri M. and Zaman, Mohd H. Mohd and Rahni, Ashrani A. Abd.},
  booktitle={IEEE-EMBS Conference on Biomedical Engineering and Sciences (IECBES)}, 
  title={Development of Real-Time Contact Force Control of a Collaborative Robot for Automated Ultrasound Scanning}, 
  year={2022},
  volume={},
  number={},
  pages={334-337},
  doi={10.1109/IECBES54088.2022.10079599}}

@INPROCEEDINGS{rtde_2,
  author={Kumar, Shitij and Arora, Sarthak and Sahin, Ferat},
  booktitle={IEEE 15th International Conference on Automation Science and Engineering (CASE)}, 
  title={Speed and Separation Monitoring using On-Robot Time-of-Flight Laser-ranging Sensor Arrays}, 
  year={2019},
  volume={},
  number={},
  pages={1684-1691},
  doi={10.1109/COASE.2019.8843326}}

@ARTICLE{log_rtde_1,
  author={Kolvig-Raun, Emil Stubbe and Kjærgaard, Mikkel Baun and Brorsen, Ralph},
  journal={IEEE Robotics and Automation Letters}, 
  title={Joint Stress Estimation and Remaining Useful Life Prediction for Collaborative Robots to Support Predictive Maintenance}, 
  year={2024},
  volume={9},
  number={4},
  pages={3554-3561},
  doi={10.1109/LRA.2024.3368296}}

@INPROCEEDINGS{log_rtde_2,
  author={Kolvig-Raun, Emil Stubbe and Kjærgaard, Mikkel Baun and Brorsen, Ralph},
  booktitle={IEEE/ACM 5th International Workshop on Robotics Software Engineering (RoSE)}, 
  title={EDDE: An Event-Driven Data Exchange to Accurately Introspect Cobot Applications}, 
  year={2023},
  volume={},
  number={},
  pages={25-30},
  doi={10.1109/RoSE59155.2023.00009}}

@INPROCEEDINGS{log_rtde_3,
  author={Kirschner, Robin Jeanne and Kurdas, Alexander and Karacan, Kübra and Junge, Philipp and Baradaran Birjandi, Seyed Ali and Mansfeld, Nico and Abdolshah, Saeed and Haddadin, Sami},
  booktitle={IEEE/RSJ International Conference on Intelligent Robots and Systems (IROS)}, 
  title={Towards a Reference Framework for Tactile Robot Performance and Safety Benchmarking}, 
  year={2021},
  volume={},
  number={},
  pages={4290-4297},
  doi={10.1109/IROS51168.2021.9636329}}

@INPROCEEDINGS{ur_rtde_1,
  author={Halim, Jayanto and Eichler, Paul and Krusche, Sebastian and Bdiwi, Mohamad and Ihlenfeldt, Steffen},
  booktitle={33rd IEEE International Conference on Robot and Human Interactive Communication (ROMAN)}, 
  title={A Hybrid Approach of No-Code Robot Programming for Agile Production: Integrating Finger-Gesture and Point Cloud}, 
  year={2024},
  volume={},
  number={},
  pages={732-739},
  doi={10.1109/RO-MAN60168.2024.10731334}}

@INPROCEEDINGS{ur_rtde_2,
  author={Hosseini, Sara and Hahn, Ingo},
  booktitle={33rd IEEE International Conference on Robot and Human Interactive Communication (ROMAN)}, 
  title={{On Experimental Energy Consumption Estimation of a 6-DoF Industrial UR3e Robot Arm Manipulator in Trajectory Planning}}, 
  year={2024},
  volume={},
  number={},
  pages={1430-1435},
  doi={10.1109/RO-MAN60168.2024.10731263}}

@INPROCEEDINGS{ur_rtde_3,
  author={Shi, Yunlei and Chen, Zhaopeng and Wu, Yansong and Henkel, Dimitri and Riedel, Sebastian and Liu, Hongxu and Feng, Qian and Zhang, Jianwei},
  booktitle={IEEE/RSJ International Conference on Intelligent Robots and Systems (IROS)}, 
  title={Combining Learning from Demonstration with Learning by Exploration to Facilitate Contact-Rich Tasks}, 
  year={2021},
  volume={},
  number={},
  pages={1062-1069},
  doi={10.1109/IROS51168.2021.9636417}}

@ARTICLE{ur_rtde_4,
  author={Shi, Yunlei and Yuan, Chengjie and Tsitos, Athanasios and Cong, Lin and Hadjar, Hamid and Chen, Zhaopeng and Zhang, Jianwei},
  journal={IEEE Transactions on Cognitive and Developmental Systems}, 
  title={A Sim-to-Real Learning-Based Framework for Contact-Rich Assembly by Utilizing CycleGAN and Force Control}, 
  year={2023},
  volume={15},
  number={4},
  pages={2144-2155},
  doi={10.1109/TCDS.2023.3237734}}

@ARTICLE{ur_rtde_5,
  author={Cheng, Zhuoqi and Zeltner, Andreas Sørensen and Årsvold, Alex Tinggaard and Schwaner, Kim Lindberg and Jensen, Pernille Tine and Savarimuthu, Thiusius Rajeeth},
  journal={IEEE Transactions on Instrumentation and Measurement}, 
  title={Active Search of Subsurface Lymph Nodes Using Robot-Assisted Electrical Impedance Scanning}, 
  year={2022},
  volume={71},
  number={},
  pages={1-11},
  doi={10.1109/TIM.2022.3147909}}

@ARTICLE{ur_rtde_6,
	author={Wu, Dianhao and Jiang, Jingang and Pan, Jie and Qian, Kun and Xue, Zhonghao},
	journal={IEEE/ASME Transactions on Mechatronics}, 
	title={{Root Canal Preparation Robot Based on Guiding Strategy for Safe Remote Therapy: System Design and Feasibility Study (2023)}}, 
	year={2025},
	volume={30},
	number={1},
	pages={84-95},
}

@INPROCEEDINGS{ur_rtde_7,
  author={Lee, Jaeyeon and Quist, Ethan and German, Stan and Kim, Michael and Fisher, Nathan},
  booktitle={19th International Conference on Ubiquitous Robots (UR)}, 
  title={Neural Network Model of eFAST Target Prediction for Robotic Ultrasound Diagnostics in Austere Environments}, 
  year={2022},
  volume={},
  number={},
  pages={153-158},
  doi={10.1109/UR55393.2022.9826265}}

@INPROCEEDINGS{ros_driver_1,
  author={Cosmin, Bucur and Alexandru, Andrei and Sorin, Tasu},
  booktitle={15th International Conference on Electronics, Computers and Artificial Intelligence (ECAI)}, 
  title={Controlling industrial robots with Simulink}, 
  year={2023},
  volume={},
  number={},
  pages={1-4},
  doi={10.1109/ECAI58194.2023.10193974}}

@INPROCEEDINGS{ros_driver_2,
  author={Moyo, Kamogelo Teddy Theodore and Luces, Jose Victorio Salazar and Ravankar, Ankit A. and Hirata, Yasuhisa and Shota, Morozumi},
  booktitle={IEEE 20th International Conference on Automation Science and Engineering (CASE)}, 
  title={Enhancing Manipulator Flexibility: Real-time Positional Control for Variable Assembly Environments using AprilTag Markers and Edge Detection}, 
  year={2024},
  volume={},
  number={},
  pages={3932-3939},
  doi={10.1109/CASE59546.2024.10711707}}

@INPROCEEDINGS{ros_driver_3,
  author={Ryu, Seungmin and Jung, Jieun and Song, Byunghun and Shin, Junho},
  booktitle={IEEE 29th International Conference on Emerging Technologies and Factory Automation (ETFA)}, 
  title={Development of Robot Applications Utilizing Asset Administration Shell for Integrated Robotics Processes}, 
  year={2024},
  volume={},
  number={},
  pages={1-4},
  doi={10.1109/ETFA61755.2024.10711037}}

@INPROCEEDINGS{ros_driver_4,
  author={Nascimento, Felipe H. N. and Cardoso, Sabrina A. and Lima, Antonio M. N. and Santos, Danilo F. S.},
  booktitle={Latin American Robotics Symposium (LARS), Brazilian Symposium on Robotics (SBR), and Workshop on Robotics in Education (WRE)}, 
  title={Synchronizing a collaborative arm’s digital twin in real-time}, 
  year={2023},
  volume={},
  number={},
  pages={230-235},
  doi={10.1109/LARS/SBR/WRE59448.2023.10333055}}

@INPROCEEDINGS{ros_driver_5,
  author={Fernando, Chandima and Olds, Daniel and Campbell, Stuart I. and Maffettone, Phillip M.},
  booktitle={IEEE International Conference on Robotics and Automation (ICRA)}, 
  title={Facile Integration of Robots into Experimental Orchestration at Scientific User Facilities}, 
  year={2024},
  volume={},
  number={},
  pages={9578-9584},
  doi={10.1109/ICRA57147.2024.10611706}}

\end{document}